\documentclass[12pt]{article}
\usepackage{amsmath}
\usepackage{amssymb}
\usepackage{times}
\usepackage{graphicx}
\usepackage{color}
\usepackage{multirow}
\usepackage[authoryear]{natbib}
\usepackage{rotating}
\usepackage{bbm}
\usepackage{latexsym}
\usepackage{appendix}

\newcommand{\captionfonts}{\normalsize}
\newcommand{\rev}{\textcolor{black}}

\makeatletter  
\long\def\@makecaption#1#2{%
  \vskip\abovecaptionskip
  \sbox\@tempboxa{{\captionfonts #1: #2}}%
  \ifdim \wd\@tempboxa >\hsize
    {\captionfonts #1: #2\par}
  \else
    \hbox to\hsize{\hfil\box\@tempboxa\hfil}%
  \fi
  \vskip\belowcaptionskip}
\makeatother   

\begin{document}
\hspace{13.9cm}1

\ \vspace{20mm}\\

{\LARGE Mirror descent of Hopfield model}

\ \\
{\bf \large Hyungjoon Soh$^{\displaystyle 1}$, Dongyeob Kim$^{\displaystyle 2}$, Juno Hwang$^{\displaystyle 1}$, Junghyo Jo$^{\displaystyle 1, \displaystyle 2, \displaystyle 3, \displaystyle 4, \displaystyle *}$}\\
{$^{\displaystyle 1}$Department of Physics Education, Seoul National University, Seoul 08826, Korea.}\\
{$^{\displaystyle 2}$Department of Physics and Astronomy, Seoul National University, Seoul 08826, Korea.}\\
{$^{\displaystyle 3}$Center for Theoretical Physics and Artificial Intelligence Institute, Seoul National University, Seoul 08826, Korea.}\\
{$^{\displaystyle 4}$School of Computational Sciences, Korea Institute for Advanced Study, Seoul 02455, Korea.}\\
{$^{\displaystyle *}$Corresponding author: jojunghyo@snu.ac.kr.}\\
%


\newpage
\thispagestyle{empty}
\markboth{}{NC instructions}
%

%
\begin{center} {\bf Abstract} \end{center}
\rev{
Mirror descent is an elegant optimization technique that leverages a dual space of parametric models to perform gradient descent. While originally developed for convex optimization, it has increasingly been applied in the field of machine learning. In this study, we propose a novel approach for utilizing mirror descent to initialize the parameters of neural networks. Specifically, we demonstrate that by using the Hopfield model as a prototype for neural networks, mirror descent can effectively train the model with significantly improved performance compared to traditional gradient descent methods that rely on random parameter initialization. Our findings highlight the potential of mirror descent as a promising initialization technique for enhancing the optimization of machine learning models.}


\section{Introduction}
Recent advancement of machine learning heavily relies on optimization techniques, since machine-learning tasks can be formulated as optimization problems in general.
The optimization of convex functions is a well-established subject~\citep{nesterov2018lectures, wright1999numerical}.
\rev{
While naive gradient-descent (GD) methods are powerful for finding minima of loss functions, natural gradient-descent (NGD) methods are more advanced as they take into account the curvature of loss functions during the optimization process. Additionally, mirror descent (MD) leverages the dual structure of parametric optimization. With these advanced optimization techniques, the insights from convex optimization are anticipated to significantly contribute to solving various machine-learning problems.
Notably, the dual structure of MD has proven to be effective in implementing NGD in machine learning without requiring explicit computation of the curvature of loss functions~\citep{azizan2021stochastic, zhan2021policy, lan2022policy, garcia2023}.
}

\rev{
In modern machine learning, deep neural networks often involve a massive number of parameters. As such, effective parameter initialization is critical for successfully training neural networks. 
Techniques such as independent component analysis~\citep{yam2002independent}, $k-$means clustering~\citep{coates2012learning}, principal component analysis~\citep{chan2015pcanet}, and linear discriminant analysis~\citep{masden2020linear} are commonly used to extract reference patterns from the data and determine initial weight parameters.
While this issue was given serious consideration in the early days of machine learning, it has become less of a priority in light of the substantial computing resources now available. 
Data-independent initialization schemes, that adjust simply the scale of randomized parameter values to prevent shrinkage or explosion during forward signal propagation through neural network layers, have become popular. Representative examples include the Xavier and He initializations~\citep{Xavier2010Understanding, He2015Delving}.
However, given the substantial energy consumption of computation~\citep{strubell2019energy}, good parameter initialization remains an important issue for reducing computational costs. Therefore, it is still an area of active research to find better techniques for initializing parameters that can reduce the computational burden~\citep{narkhede2022review}.
}

\rev{
In this study, we find compelling evidence that MD is naturally linked to data-driven parameter initialization in parametric models that belong to the exponential family. Specifically, we observe that the dual variable of MD corresponds to the expectation value of the empirical data distribution, implying that the initial parameter value for the dual space is naturally related to data. 
}

\rev{
This paper is organized as follows. In Section~\ref{MD}, we review the mirror descent to convey its essence for subsequent discussion.
In Section~\ref{Hopfield}, we adopt the Hopfield model as an ideal prototype model of neural networks to implement our idea. In Section~\ref{Result}, by utilizing MD to optimize the Hopfield model, we demonstrate that MD outperforms usual optimization methods such as GD and NGD with random parameter initialization. Finally, we conclude our findings in Section~\ref{Conclusion}. To ensure the paper is self-contained, we briefly introduce GD and NGD in the Appendix.
}

\section{\rev{Mirror descent}}
\label{MD}

\rev{
GD can be interpreted as a solution of the following proximal optimization,
\begin{equation}
\label{eq:proxi2}
    \theta^{t+1} = \operatorname*{argmin}_\theta \bigg( L(\theta) + \frac{1}{\alpha} D(\theta, \theta^t) \bigg),
\end{equation}
where $D(\theta, \theta^t) = || \theta - \theta^t ||^2$ is the squared Euclidean distance.
Then, GD iteratively updates the parameter $\theta^{t+1}$ by minimizing the augmented loss function that includes a proximal constraint, which ensures that $\theta^{t+1}$ remains ``close'' to $\theta^{t}$. 
Here, the distance between $\theta$ and $\theta^t$ can be generally defined as the Bregman divergence:
\begin{equation}
\label{eq:bregman}
    D(\theta, \theta^t) = F(\theta) - F(\theta^t) - \nabla_\theta F(\theta^t) \cdot (\theta - \theta^t)
\end{equation}
for a convex function of $F(\theta)$. 
The Bregman divergence is a measure of the difference between the function $F(\theta)$ and its first-order Taylor approximation, $F(\theta^t) + \nabla_\theta F(\theta^t) \cdot (\theta - \theta^t)$, where the divergence vanishes only at $\theta = \theta^t$.
Now, the optimal solution of Eq.~(\ref{eq:proxi2}) satisfies 
\begin{equation}
\label{eq:optim}
    \nabla_\theta L(\theta^t) + \frac{1}{\alpha} \bigg(\nabla_\theta F(\theta^{t+1}) - \nabla_\theta F(\theta^{t}) \bigg) = 0,
\end{equation}
which is derived using the first-order Taylor approximation of the loss function $L(\theta)$ around $\theta^t$.
}

\begin{figure}[t]
\hfill
\begin{center}
\includegraphics[width=4.5in]{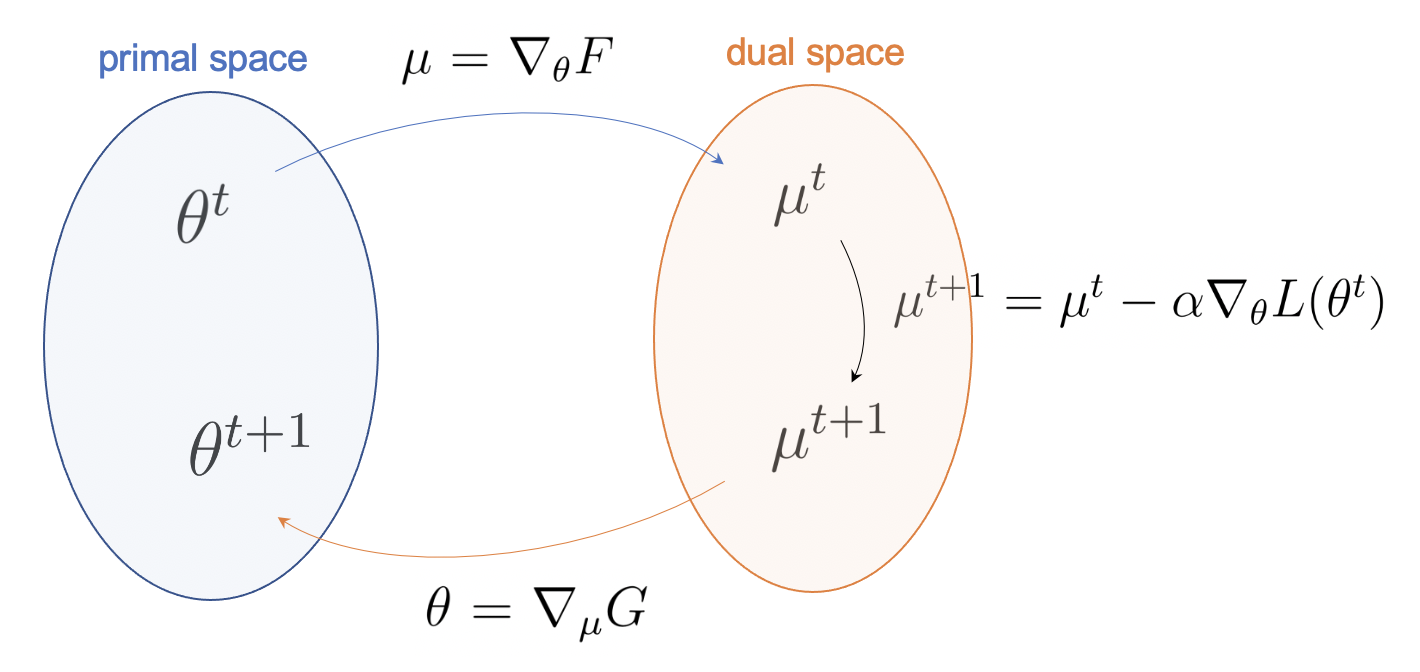}
\end{center}
\caption{Mirror descent. The algorithm consists of three steps: (i) transformation from the natural parameter $\theta^t$ in primal space to $\mu^t$ in dual space; (ii) update of $\mu^t \to \mu^{t+1}$; and (iii) inverse transformation from $\mu^{t+1}$ in dual space to $\theta^{t+1}$ in primal space.}
\label{Fig:duality}
\end{figure}

To make this condition more readable, we can introduce a new variable, $\mu \equiv \nabla_\theta F$, which is conjugate to $\theta$.
Since the convex function $F(\theta)$ has distinct slopes $\mu$ for every position $\theta$, knowing $\mu$ is equivalent to knowing $\theta$.
Then, the optimal condition of Eq.~(\ref{eq:optim}) can be reformulated as follows,
\begin{equation}
\label{eq:gd2}
    \mu^{t+1} = \mu^t - \alpha \nabla_\theta L(\theta^t).
\end{equation}
At first sight, this equation looks like the GD formula.
However, note that the gradient $\nabla_\theta L$ is computed for the change of $\theta$, not for the change of $\mu$.
It is remarkable that Eq.~(\ref{eq:gd2}) corresponds to NGD for the update in $\mu$ space, although it looks like GD~\citep{raskutti2015information}.
To explicitly explain this point, another convex function $G(\mu)$ is introduced as a Legendre transformation of the original convex function $F(\theta)$: 
\begin{equation}
    F(\theta) + G(\mu) = \theta \cdot \mu.
\end{equation}
The duality between $\theta$ and $\mu$ imposes $\mu = \nabla_\theta F$ and $\theta = \nabla_\mu G$.
Then, we are ready to interpret $\nabla_\theta L$ for the change of $\mu$ using chain rules,
\begin{equation}
    \frac{\partial L}{\partial \theta_i} = \sum_j \frac{\partial \mu_j}{\partial \theta_i} \frac{\partial L}{\partial \mu_j} = \sum_j \bigg[ \frac{\partial^2 G}{\partial \mu_i \partial \mu_j} \bigg]^{-1}\frac{\partial L}{\partial \mu_j},
\end{equation}
where $\theta_i = \partial G/\partial \mu_i$ is used.
Thus, GD-like Eq.~(\ref{eq:gd2}) implies NGD:
\begin{equation}
    \mu_i^{t+1} = \mu_i^t - \alpha \sum_j \bigg[ \frac{\partial^2 G}{\partial \mu_i \partial \mu_j} \bigg]^{-1}\frac{\partial L(\mu^t)}{\partial \mu_j}.
\end{equation}
Summarizing the updating procedures of MD as shown in Figure~\ref{Fig:duality}, (i) current $\theta^t$ in primal space is transformed to $\mu^t$ in dual space.
(ii) Then, it is updated to $\mu^{t+1}$ through the first-order GD. (iii) Finally, the updated $\mu^{t+1}$ is transformed back to $\theta^{t+1}$ in primary space.

MD has a great computational merit that the first-order GD in dual space implicitly realizes the second-order NGD.
However, one big obstacle for using MD is the inverse transformation to the primal space, which is usually not available. One exceptional case is using the $p$-norm as $F(\theta) = (\sum_i |\theta_i|^p)^{1/p}$. This setup allows easy transformations between primal and dual spaces because the dual function $G(\mu) = (\sum_i |\mu_i|^q)^{1/q}$ corresponds to the $q$-norm given $1/p + 1/q=1$~\citep{gentile1999robustness}. The MD with $p$-norm has been nicely applied in deep learning~\citep{azizan2021stochastic}.

\rev{
Another useful application of MD is in policy updates for reinforcement learning~\citep{lan2022policy, zhan2021policy}. Specifically, when the parameter vector $\theta$ has a normalization constraint, such as $\sum_i \theta_i = 1$ as a probability distribution, and the Bregman divergence $D(\theta, \theta')= \sum_i \theta_i \log (\theta_i/\theta'_i)$ is the Kullback-Leibler divergence, the solution to Eq.~(\ref{eq:proxi2}) provides the subgradient algorithm with nonlinear projections, as shown by \cite{beck2003mirror}.}


\rev{
It is important to note that both the $p$-norm and subgradient algorithms rely on a data-independent distance $D(\theta, \theta')$ between models. However, for machine learning problems, it is often desirable to use a data-dependent distance $D(\theta, \theta')$ between models, where the model function $F(\theta)$ is dependent on data. Therefore, our goal is to implement the optimization procedure of MD into a more realistic setting for machine learning problems by incorporating a data-dependent distance.
}

\section{Hopfield model}
\label{Hopfield}
\rev{The Hopfield model can serve as an excellent prototype for applying MD with a data-dependent manner.}
It was developed to explain associative memory in neuroscience~\citep{hopfield1982neural}, and later it works as a basic model for generative models of neural networks.
Suppose we observe a firing pattern of $n$ neurons, $x = (x_1, x_2, \cdots, x_n)$, and obtain a data of $M$ patterns, $\{ x(t) \}_{t=1}^M $, where $t$ is not necessary to be time, but it may be a sample index.
Given the $M$ states, one can estimate the relative frequency of them as $\hat{P}(x) = 1/M \sum_{t=1}^M \delta(x - x(t))$. To model the empirical distribution, Hopfield proposed an exponential-family model:
\begin{equation}
\label{eq:hopfield}
    P(x; W, b) = \frac{\exp(b\cdot x + x W x^T)}{Z},
\end{equation}
where $Z = \sum_x \exp(b\cdot x + x W x^T)$ is a normalization factor, called partition function in statistical mechanics. Note that $x^T$ represents transpose of $x$. Here the model parameter $b_i$ represents the bias of the $i$-th neuron determining its average activity, and $W_{jk}$ represents the coupling strength between the $j$-th and the $k$-th neurons.
Then, Hopfield noticed a good choice of parameter values:
\begin{align}
\label{eq:hopfieldsol}
    b_i &= \mathbb{E}_{\hat{P}}[x_i], \nonumber \\
    W_{jk} &= \mathbb{E}_{\hat{P}} [x_j x_k],
\end{align}
where $\mathbb{E}_{\hat{P}}[O] \equiv \sum_x O(x) \hat{P}(x)$ denotes expectation calculation. 
It is noted that the coupling strength $W_{jk}$ depends on the correlation $\mathbb{E}_{\hat{P}} [x_j x_k]$ between two neurons. This is consistent with the Hebb's rule, `fire together and wire together'~\citep{hebb2005organization}. 
Introducing a parameter vector $\theta \equiv \{b_1, b_2, \cdots, W_{12}, \cdots\}$ and an operator vector $O \equiv \{ x_1, x_2, \cdots, x_1 x_2, \cdots\}$, we can simplify the model  of Eq.~(\ref{eq:hopfield}) and the Hopfield solution of Eq.~(\ref{eq:hopfieldsol}) as
\begin{equation}
    P(x; \theta) = \frac{\exp(\theta \cdot O(x))}{Z}, \phantom{MM}\theta_i = \mathbb{E}_{\hat{P}} [O_i].
\end{equation}
This Hopfield solution makes sense because the probability of a pattern $x$ becomes large once $O(x)$ aligns with the average pattern of $\theta = \mathbb{E}_{\hat{P}} [O]$.

Indeed one can have a better solution that makes the model distribution $P(x; \theta)$ closer to the data distribution $\hat{P}(x)$~\citep{ackley1985learning}.
Boltzmann machine defines the distance between the two distributions using the Kullback-Leibler divergence:
\begin{equation}
    L(\theta) = D_{KL} \bigg[ \hat{P}(x) || P(x; \theta) \bigg] = \sum_x \hat{P}(x) \ln \frac{\hat{P}(x)}{P(x; \theta)}.
\end{equation}
Given $L(\theta)$, it is straightforward to obtain its gradient:
\begin{equation}
\label{eq:gradient}
\nabla_\theta L = \mathbb{E}_P[O] - \mathbb{E}_{\hat{P}}[O].
\end{equation}
Then, we can minimizes the distance by using GD with $\nabla_\theta L$.
Once the model and data expectations of $\mathbb{E}_P[O]$ and $\mathbb{E}_{\hat{P}}[O]$ are consistent, the gradient vanishes. 
Here we will use MD instead of this GD.

First, we start from the partition function, $Z = \sum_x \exp(\theta \cdot O(x))$. Its logarithm becomes a cumulant generating function, $F \equiv \ln Z$.
One can easily obtain cumulants by differentiating $F$. For example, the first cumulant is obtained by
\begin{equation}
    \nabla_\theta F = \sum_x O(x) \frac{\exp(\theta \cdot O)}{Z} = \mathbb{E}_P[O] = \mu.
\end{equation}
It is the unique feature of the exponential family that the conjugate variable $\mu$ corresponds to the expectation value. 
Now we apply MD for the Hopfield model in the exponential family:
\begin{itemize}
    \item[(i)] Transformation to dual space by computing expectation, $\mu^t = \mathbb{E}_{P(x;\theta^t)}[O]$;
    \item[(ii)] Parameter update through GD, $\mu^{t+1} = \mu^t - \alpha \nabla_\theta L(\theta^t)$;
    \item[(iii)] Reverse transformation to primary space, $\theta^{t+1} = \nabla_\mu G(\mu^{t+1})$.
\end{itemize}
Step (i) is computed by virtue of the exponential-family model.
Then, step (ii) is also well defined with the gradient of Eq.~(\ref{eq:gradient}).
As always, the obstacle is in the reverse transformation of step (iii),
because we do not have the explicit function $G(\mu)$, the Legendre transformation of $F(\theta)$. 


We solve this problem by using the Taylor expansion of $G(\mu^{})$ at around $\mu = \mu^{t}$.
\begin{equation}
    G(\mu^{}) \approx G(\mu^t) + \sum_i \frac{\partial G(\mu^t)}{\partial \mu_i} (\mu_i^{} - \mu_i^t) + \frac{1}{2} \sum_{j,k} \frac{\partial^2 G(\mu^t)}{\partial \mu_j \partial \mu_k} (\mu_j^{} - \mu_j^t)(\mu_k^{} - \mu_k^t).
\end{equation}
For an infinitesimal change of $\mu^t \rightarrow \mu^{t+1}$, it is sufficient to know the local landscape of $G(\mu)$.
By differentiating $G(\mu)$ with respect to $\mu_i$ and evaluating at $\mu = \mu^{t+1}$, we have
\begin{equation}
\label{eq:return1}
    \frac{\partial G(\mu^{t+1})}{\partial \mu_i} = \frac{\partial G(\mu^{t})}{\partial \mu_i} + \sum_j \frac{\partial^2 G(\mu^t)}{\partial \mu_i \partial \mu_j} (\mu_j^{t+1} - \mu_j^t),
\end{equation}
where the curvature is
\begin{equation}
    \frac{\partial^2 G(\mu^t)}{\partial \mu_i \partial \mu_j} = \frac{\partial \theta_i}{\partial \mu_j} = \bigg[ \frac{\partial \mu_j}{\partial \theta_i} \bigg]^{-1} = \bigg[ \frac{\partial^2 F(\theta^t)}{\partial \theta_i \partial \theta_j} \bigg]^{-1}.
\end{equation}
Here $\partial \mu_j / \partial \theta_i$ is nothing but the covariance matrix of operators $O$,
\begin{align}
\label{eq:covariance}
    \frac{\partial \mu_j}{\partial \theta_i} &= \frac{\partial}{\partial \theta_i} \sum_x O_j(x) \frac{\exp(\theta \cdot O(x))}{Z} \nonumber \\
    &= \mathbb{E}_{P}[O_i O_j] - \mathbb{E}_{P}[O_i] \mathbb{E}_{P}[O_j] = C_{ij}.
\end{align}
Then, Eq.~(\ref{eq:return1}) becomes
\begin{equation}
\label{eq:return2}
    \theta_i^{t+1} = \theta_i^t + \sum_j \big[ C^{-1} \big]_{ij} (\mu_j^{t+1} - \mu_j^t). 
\end{equation}
This allows us to return primary space.
One caveat of this process is that we need to compute the curvature or the covariance matrix.
This process \rev{loses} the merit of MD that implies the second-order method without explicit calculation of the curvature.
When the update step (ii) is incorporated into Eq.~(\ref{eq:return2}), NGD emerges as
\begin{equation}
    \theta_i^{t+1} = \theta_i^t + \alpha \sum_j \bigg[\frac{\partial^2 F(\theta^t)}{\partial \theta_i \partial \theta_j} \bigg]^{-1} \frac{\partial L(\theta^t)}{\partial \theta_j}.
\end{equation}
For the Hopfield model, the covariance matrix is identical to the Hessian matrix:
\begin{equation}
    C_{ij} = \frac{\partial^2 F(\theta^t)}{\partial \theta_i \partial \theta_j} = \frac{\partial^2 L(\theta^t)}{\partial \theta_i \partial \theta_j}.
\end{equation}

In summary, by using the Taylor expansion, we can run the MD algorithm for the Hopfield model. Furthermore, we realized that the MD update becomes identical to NGD.
Then, one may ask what MD and NGD are different.
An important difference exists between the abstract $\theta$ in primary space and the observable $\mu$ in dual space.
For GD or NGD, one should start from a random $\theta^0$ due to no prior knowledge on $\theta$.
In contrast, $\mu^0$ in dual space is different.
The dual parameter of the exponential family corresponds to an expectation value $\mu = \mathbb{E}_P[O]$. 
Then, we are tempted to adopt the Hopfield solution as an initial value of $\mu^0 = \mathbb{E}_{\hat{P}}[O]$, since the data distribution is the most confident distribution at the moment.
Note that we still do not know $\theta^0$ that is conjugate to $\mu^0 = \mathbb{E}_{P(x; \theta^0)}[O]= \mathbb{E}_{\hat{P}}[O]$.

Starting from $\mu^0$ in dual space looks great, but it has a fundamental problem. Since $\mu^0$ represents the data distribution, it implies that $L(\theta^0) = D_{KL}[\hat{P}(x)||P(x;\theta^0)]=0$ and $\nabla_\theta L(\theta^0) = 0$.
The vanishing gradient keeps $\mu^1 = \mu^0$ from updating.
This result makes a perfect sense, since the data distribution is the most confident one with no need to change.
As an alternative choice of $\mu^0 = \mathbb{E}_{\hat{P}}[O]$, we consider the Hopfield solution as an initial value of $\theta^0$ instead of $\mu^0$.
\rev{It should be noted that the $\mu^0$, conjugate to this choice of $\theta^0= \mathbb{E}_{\hat{P}}[O]$, is no longer $\mu^0 = \mathbb{E}_{\hat{P}}[O]$, although $\mu^0$ may be close to $\mathbb{E}_{\hat{P}}[O]$ because the Hopfield solution makes $P(x; \theta^0)$ pretty close to $\hat{P}(x)$.
}
Given the choice of data-driven $\theta^0$, MD can be interpreted as the NGD starting from the Hopfield solution.


\begin{figure}
\hfill
\begin{center}
\includegraphics[width=5.5in]{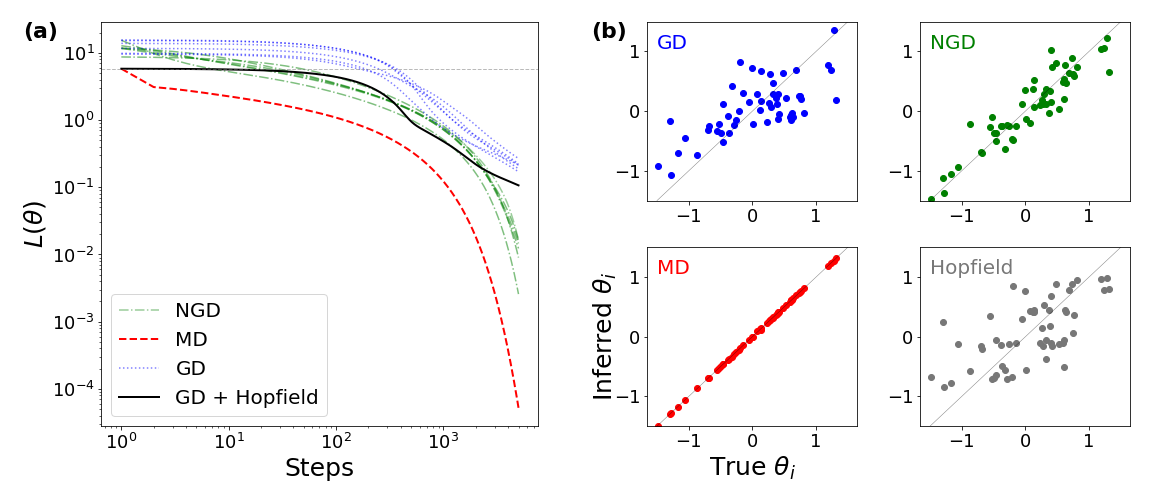}
\end{center}
\caption{(a) Learning performances of different gradient methods. The Hopfield model $P(x; \theta)$ learns the data distribution $\hat{P}(x)$ of $x = (x_1, x_2, \cdots, x_{10})$. The discrepancy between model and data distributions is defined as a loss $L(\theta)$. Plotted are the loss changes of (i) gradient descent (GD) with different random parameter initializations (blue dotted lines); (ii) natural gradient descent (NGD) with different random parameter initializations (green dash-single dotted lines); and (iii) GD+Hopfield, starting from the Hopfield solution (black solid line); and (iv)  mirror descent (MD), starting from the Hopfield solution (red dashed line). For the simulation, a learning rate $\alpha = 0.001$ is used \rev{for GD, NGD, and MD}.
(b) Accuracy of inference. The parameter of the Hopfield model is inferred from (i) GD (black); (ii) NGD (green); (iii) MD (red); and (iv) the Hopfield solution (grey).
True $\theta_i$ is sampled from the standard normal distribution $\mathcal{N}(0,1)$.
}
\label{Fig:loss}
\end{figure}

\section{Results}
\label{Result}
Now we examine the performance of MD applied in the Hopfield model by comparing the performance of the usual GD algorithm.
To conduct this experiment, we first need to synthesize data from the Hopfield model.
We obtain a true parameter set of the Hopfield model by sampling from the standard normal distribution as $\theta_i \sim \mathcal{N}(0, 1)$. 
Given the model parameter $\theta$, we sampled $n$-dimensional binary states $x = (x_1, x_2, \cdots, x_n)$ from the model distribution $P(x; \theta)$, and obtained their data distribution $\hat{P}(x)$.
The goal is to infer the true $\theta$ by just using the data distribution.
To circumvent the uncertainty originated from a finite size of samples, we assume that we have infinite samples. 
\rev{
Given an infinite number of samples, the estimated distribution $\hat{P}(x)$ should be identical to the ideal distribution $P(x; \theta)$. Therefore, we use $P(x; \theta)$ as a substitute for $\hat{P}(x)$ with infinite samples. Additionally, to make it possible to sum probability distributions over the full configuration space in a tractable way, we limit the state dimension of the binary variable $x$ to a value $n < 20$. For the purpose of the following experiment, we chose $n = 10$, without loss of generality.
}

We compare the convergence speed and the inference accuracy of different gradient descent methods: (i) GD; (ii) GD+Hopfield starting from the Hopfield solution; (iii) NGD; and (iv) MD.
It is expected that the initial loss of MD and GD+Hopfield is lower than the loss of GD and NGD because starting from the Hopfield solution uses prior information of data unlike random initial starting.
Furthermore, the second-order methods of MD and NGD lower the loss more dramatically during early iterations (Figure~\ref{Fig:loss}a).
Note that actual computation time is not beneficial because the computation of the curvature and its inversion requires additional computation.
After sufficient iterations, we examined the accuracy of inferred $\theta$, compared with true $\theta$, and confirmed that MD provides the most accurate inference (Figure~\ref{Fig:loss}b).

The inversion of covariance matrix $C$ has no problems for a small dimension $n$ and small values of $\theta$.
However, when $n$ and/or $\theta$ get larger, the inversion starts to be fragile with poor inference.
To make the inversion robust, we modified the covariance matrix as $\tilde{C} = C + \epsilon I$, where $I$ is an identity matrix.
Nevertheless, NGD that starts from random initial $\theta^0$ is still very unstable, because some eigenvalues $\lambda$ of $\tilde{C}^{-1}$ are very large to make the update of $\theta^{t+1}$ divergent.
Then, we frequently observed that the loss of NGD easily diverges. 
The divergence of loss can be suppressed by using a smaller learning rate (i.e., inverse of the largest $\lambda$).
The addition of $\epsilon I$ regularizes the largest $\lambda$ to be bounded to an order of $1/\epsilon$. 
Then, the update of $\theta^{t+1}$ stays finite for $\alpha/\epsilon < 1$.
Unlike the curvature evaluated at the random initial $\theta^0$, the curvature at the Hopfield solution of $\theta^0= \mathbb{E}_{\hat{P}}[O]$, shows quite robust inversion of $\tilde{C}$.
This allows the stable update of $\mu^{t+1}$ even with a marginal learning rate (e.g., $\alpha \sim 0.1$).


\begin{figure}
\hfill
\begin{center}
\includegraphics[width=3.5in]{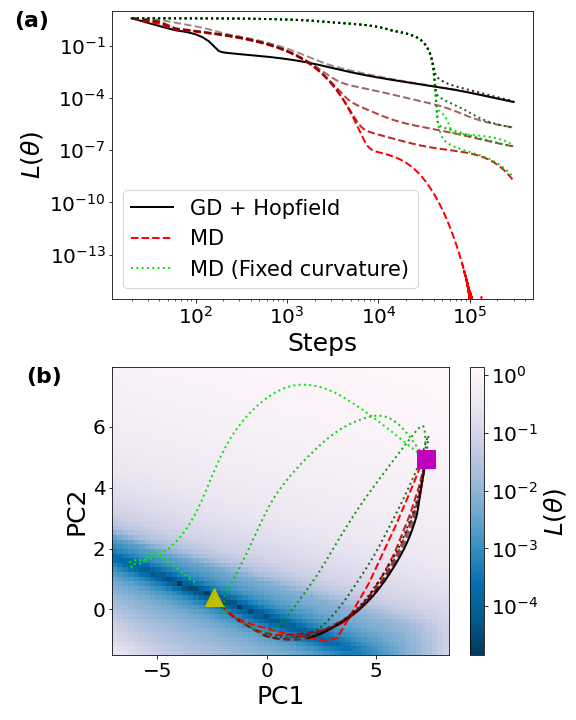}
\end{center}
\caption{Mirror descent with a fixed curvature. (a) Loss curves of (i) gradient descent (GD) starting from the Hopfield solution with a learning rate $\alpha=0.1$ (black solid line), (ii) mirror descent (MD) with updating curvatures and $\alpha=0.001$ and $\epsilon=[10^{-6}, 10^{-5}, 10^{-4}, 10^{-3}, 10^{-2}]$ (\rev{With increasing $\epsilon$ values, the six dashed curves move closer to the black solid line, and their colors change from bright red to dark red}), and (iii) MD with a fixed curvature and $\alpha=0.001$ and $\epsilon=[10^{-6}, 10^{-5}, 10^{-4}, 10^{-3}, 10^{-2}]$ (\rev{With increasing $\epsilon$ values, the six dotted curves move closer to the black solid line, and their colors change from bright green to dark green}). (b) Learning paths of parameters are projected on the principal component analysis (PCA) plane. Each path departs from the Hopfield solution (top right square), and approaches into a global optimum (bottom left triangle). True $\theta_i$ is sampled from $\mathcal{N}(0, 1.5)$.
}
\label{Fig:hopfield_curvature}
\end{figure}

MD and NGD show more effective parameter updates because they use the curvature information of the loss.
The curvature is evaluated at $P(x; \theta)$ with Eq.~(\ref{eq:covariance}) in MD. 
This means that the curvature should be obtained from the updated model parameters for every iteration. 
However, to reduce the computational cost, one may consider to use a fixed curvature $C^0$ evaluated at the data distribution $\hat{P}(x)$ as a proxy of updating curvatures at every step. 
The loss curve for these two schemes are compared in Figure~\ref{Fig:hopfield_curvature}(a). Overall, MD with updating curvatures (red dashed lines) converges faster than MD with the fixed curvature (green dotted lines). However, both MD methods outperform GD (black solid line) in spite of the 100 times smaller learning rate compared to the learning rate for GD.

Indeed the fixed curvature leads to different updating directions due to the discrepancy from the actual curvature evaluated at the updating $\theta^t$. 
This `detouring' takes extra steps toward a global optimum. 
To visualize the different learning paths on a two-dimensional space, we adopt the principle component analysis (PCA).
We first makes a contour map of $L(\theta)$ for $\theta$ estimated at different (PC1, PC2).
Then, we put the learning paths of GD, MD with updating curvatures, and MD with the fixed curvature (Figure~\ref{Fig:hopfield_curvature}(b)).
Every path starts at the Hopfield solution (top right white circle), and approaches into a single global minimum (bottom left red circle) of the exponential-family model.
Due to the information loss under the dimension reduction of PCA, some overshoot is observed in the learning paths on the PCA plane.
We demonstrated two important points.
First, MD with the fixed curvature shows a large detour to the global minimum.
Nevertheless, once it approaches to the minimum, the fixed curvature becomes consistent with the curvature evaluated at the global minimum because the model distribution approaches into the data distribution.
Second, increasing the regularization hyperparameter $\epsilon$ allows to approximate the curvature matrix $\tilde{C} = C + \epsilon I$ as an identity matrix.
Then, MD approaches to GD. Indeed, as we increase $\epsilon$, the learning path approaches to the path of GD.

\section{Conclusion}
\label{Conclusion}
\rev{
Our study reveals an intriguing connection between the mirror descent and the Hopfield solution. The mirror descent offers an elegant formalism for achieving natural gradient descent without explicitly calculating the curvature of loss functions, through the use of dual structure in parametric models. Our contribution lies in recognizing the relationship between the conjugate variables of exponential family models in dual space and the expectation values of empirical data distributions. We focused specifically on the Hopfield model, which is a well-known prototype of neural networks falling under the exponential family models category. The solution to the Hopfield model parameters is proposed as the mean and correlations of empirical data distributions. Through the application of mirror descent, we can explain the Hopfield solution as conjugate variables in dual space, allowing for data-driven parameter initialization.
}

The data-driven parameter initialization is an important subject in machine learning~\citep{seuret2017pca, das2021data, chumachenko2022feedforward}.
We emphasize that our approach is fundamentally different from the previous popular approaches, such as Xavier and He initialization~\citep{Xavier2010Understanding, He2015Delving}, that adjust initial scales of parameters depending on network size. Those approaches do not incorporate any information from data.
Furthermore, the mirror descent enjoys the second-order method that considers the loss curvature of updating model parameters.
In summary, the mirror descent can provide a nice optimization scheme with data-driven parameter initialization and update for training neural networks.


\rev{
In a recent study by \cite{cachi2020fast}, it was demonstrated that using input data as initial weights in spiking neural networks results in a tenfold reduction in training time without sacrificing accuracy. This finding suggests that our idea has potential beyond proof of concept. However, for practical machine learning models, 
several challenges must be addressed before the general application of our approach. These include the incorporation of hidden variables that do not appear in data and the generalization of the idea of data-driven parameter initialization beyond exponential families. As such, while our approach shows promise, further work is required to overcome these challenges and fully realize its potential in practical machine learning applications.
}

\subsection*{Acknowledgments}
This work was supported in part by the Creative-Pioneering Researchers Program through Seoul National University, the National Research Foundation of Korea (NRF) grant (Grant No. 2022R1A2C1006871) (J. J.), and the Summer Internship Program of Artificial Intelligence Institute of Seoul National University (D. K.).


\appendix
\subsection*{\rev{Appendix: Natural gradient-descent method}}
\label{Appendix}
Optimization problems can be formulated to find a minimum or maximum point $\theta^*$ of a designed objective function $L(\theta)$.
Note that the model parameter $\theta = (\theta_1, \theta_2, \cdots)$ is a vector in general.
Here, if $L(\theta)$ is a convex function, iterative updates of $\theta$,
\begin{equation}
\label{eq:gd}
    \theta_i^{t+1} = \theta_i^t - \alpha \frac{\partial L(\theta^t)}{\partial \theta_i},
\end{equation}
that follow the steepest direction of $L(\theta)$, guarantee to reach $\theta^*$.
The learning rate $\alpha$ controls the degree of updates.
This is the gradient-descent (GD) method.
This method has some unsatisfactory points.
First, GD does not say anything about a starting position of $\theta^0$.
Second, the learning rate is homogeneous ($\alpha_i = \alpha$) for every direction, which is not necessary.
Third, the intuitive formula of Eq.~(\ref{eq:gd}) is not {\it covariant} under a transformation of $\theta$. Suppose that we doubles $\theta_i \rightarrow 2 \theta_i$. Then, $\theta_i^{t+1}$ and $\theta_i^t$ doubles, but $\partial L/\partial \theta_i$ halves.

To make the formula covariant, the natural gradient-descent (NGD) method is developed:
\begin{equation}
    \theta_i^{t+1} = \theta_i^t - \alpha \sum_j \bigg[\frac{\partial^2 L(\theta^t)}{\partial \theta_i \partial \theta_j}  \bigg]^{-1} \frac{\partial L(\theta^t)}{\partial \theta_j},
\end{equation}
where the Hessian matrix, or the curvature, of $L(\theta)$ is considered for more effective updates.
\rev{
It is important to note that in order to make NGD covariant, it is not necessary for the curvature matrix to be the Hessian matrix. For instance, the Fisher information matrix is another option that can be used.
}
NGD is a second-order method that requires to compute the Hessian matrix and its inverse.
This additional computation is much costly for updating high-dimensional $\theta$.
For practical implementation, many smart numerical techniques, such as Broyden-Fletcher-Goldfarb-Shanno (BFGS)~\citep{davidon1991variable}, have been developed to approximate the inverse of the Hessian matrix.
NGD has also been implemented for deep learning with Kronecker-factored approximate curvature (K-FAC)~\citep{martens2015optimizing}.
\rev{
Recently, a method called FISHLEG (Garcia 2023) has been developed to address the inversion of the Fisher information matrix using Legendre-Fenchel duality.}

\bibliographystyle{APA}
\bibliography{reference}




\end{document}